\documentclass{article}

\usepackage{graphics} % for pdf, bitmapped graphics files
\usepackage{graphicx} 
\usepackage{epsfig} % for postscript graphics files
\usepackage{mathptmx} % assumes new font selection scheme installed
\usepackage{times} % assumes new font selection scheme installed
\usepackage{amsmath} % assumes amsmath package installed
\usepackage{amssymb}  % assumes amsmath package installed
\usepackage{cite}
\usepackage{algorithm}
\usepackage[noend]{algpseudocode}
\usepackage{verbatim}
\usepackage{commath}
\usepackage{bm}
\usepackage{booktabs}
\usepackage{multirow}
\usepackage{caption}
\usepackage[final]{corl_2018} % Uncomment for the camera-ready ``final'' version

\newcommand{\squishlist}{
\begin{list}{${\bullet}$}
	{ \setlength{\itemsep}{0pt}
		\setlength{\parsep}{3pt}
		\setlength{\topsep}{3pt}
		\setlength{\partopsep}{0pt}
		\setlength{\leftmargin}{1.5em}
		\setlength{\labelwidth}{1em}
		\setlength{\labelsep}{0.5em} } }
\newcommand{\squishend}{
\end{list}  
}

\usepackage[font={small,bf}]{caption}

\usepackage{titlesec}

\usepackage{etoolbox}

\makeatletter
\patchcmd{\ttlh@hang}{\parindent\z@}{\parindent\z@\leavevmode}{}{}
\patchcmd{\ttlh@hang}{\noindent}{}{}{}
\makeatother

\titlespacing{\section}{0pt}{*0}{*0}
\titlespacing{\subsection}{0pt}{*0}{*0}
\titlespacing{\subsubsection}{0pt}{*0}{*0}

\title{Inferring geometric constraints in human demonstrations}

\author{
  Guru Subramani\\
  Department of Mechanical Engineering\\
  University of Wisconsin - Madison\\
  United States\\
  \texttt{gsubramani@wisc.edu} \\
  \And
  Michael Zinn \\
  Department of Mechanical Engineering \\
  University of Wisconsin - Madison \\
  United States\\
  \texttt{mzinn@wisc.edu} \\
  \AND
  Michael Gleicher \\
  Department of Computer Sciences \\
  University of Wisconsin - Madison \\
  United States\\
  \texttt{gleicher@cs.wisc.edu}
}

\begin{document}
\maketitle

%===============================================================================

\begin{abstract}
% abstract
This paper presents an approach for inferring geometric constraints in human demonstrations. In our method, geometric constraint models are built to create representations of kinematic constraints such as \textit{fixed point}, \textit{axial rotation}, \textit{prismatic motion}, \textit{planar motion} and  others across multiple degrees of freedom. Our method infers geometric constraints using both kinematic and force/torque information. The approach first fits all the constraint models using kinematic information and evaluates them individually using position, force and moment criteria. Our approach does not require information about the constraint type or contact geometry; it can determine both simultaneously. We present experimental evaluations using instrumented tongs that show how constraints can be robustly inferred in recordings of human demonstrations.
\end{abstract}


%===============================================================================

%Introduction
\section{Introduction}
Geometric constraints are a key part of physical tasks. These constraints impose restrictions on how an object may physically move in the environment. Understanding the geometry of these constraints makes it easier to interact with them effectively. For example, opening a door is easier if one knows that the door is attached to a hinge and where the axis of the hinge is. 
This paper introduces a method for inferring geometric constraints from recorded human demonstrations of people performing tasks. Our method automatically identifies the types of constraints and their parameters robustly by incorporating both kinematic and force/moment information. For example, given a recording of a person opening a door, our method can identify that there is a hinge constraint and determine the location and axis of the hinge.

Knowledge of constraints can be useful in robotics applications. It allows robots to be programmed to use hybrid force-position control to interact with the constraint. For example, to erase a whiteboard, one applies pressure against the plane while moving long it. Such constraints are typically specified manually; constraint inference can simplify the programming process. As part of programming by demonstration, explicit inference of constraints offers the potential to allow the parameters of hybrid force-position control to be inferred from demonstrations. 
%Inferred constraints also offer causal information about demonstrated movements, enabling better generalization.  
However, existing approaches to inferring constraints are limited.   

\begin{figure}[t]
	\includegraphics[width=5.5in]{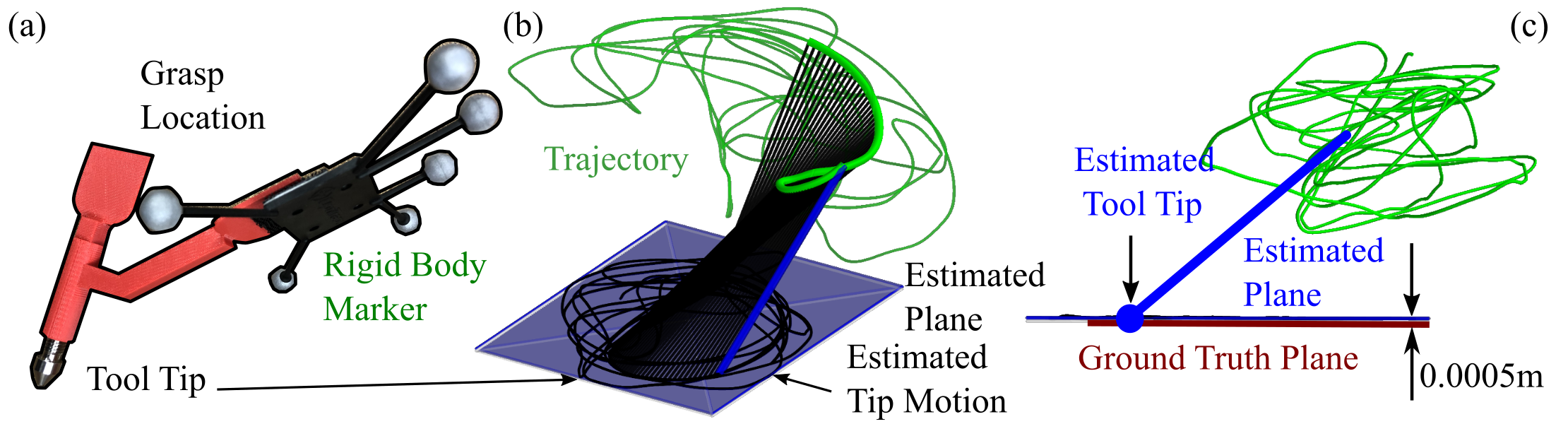}
	\caption{\textbf{(a)} Stylus (Rigid body) – A human demonstrator moves the stylus (using instrumented tongs shown in figure (\ref{fig:tongs}) with its tip against a planar surface to estimate the planar surface geometry and tool tip location. \textbf{(b)} shows the trajectory of the rigid body in green. After the demonstration, our algorithm estimates both the location of the plane in the global coordinate frame and the tool tip position relative to the rigid body frame. \textbf{(c)} The estimated plane location compared to ground truth is within the tolerance (sub-mm) of the motion capture system.}
	\vspace*{-0.2in}
	\label{fig:teaser}
\end{figure}

Our constraint inference method can model and infer complicated constrained motions with multiple degrees of freedom(DOF) such as the constrained motion of a pen when it draws on paper, without any knowledge of the pen's geometry. 
As shown in figure \ref{fig:teaser}, the tip location is estimated along with the parameters of the planar surface. Our method incorporates forces and moments to distinguish between constraints that are ambiguous when only kinematic information is used. For instance, our method can distinguish between the arc motion of a hinge constraint and circular motion in free space by incorporating forces and moments. 

The key insights behind our method are (1) the constrained motion of an object can be modeled as geometric constraints on a rigid body, (2) kinematic information(position and orientation) alone is insufficient to determine a constraint type and (3) force/moment information is useful to distinguish between constraint types. 
Therefore, our approach takes as input not only positions and orientations, but also the applied forces and moments of the tool performing the task. 

Our method can identify a wide range of standard geometric constraints and their geometric parameters without requiring knowledge of the detailed geometry of the tool. Prior methods cannot model, identify and distinguish between the variety of constraint models we consider and do not use their physical properties such as incorporating reaction forces and moments, therefore limiting their robustness with which they can identify these constraints (e.g. \citep{perez2017c}, \citep{Niekum2015}, \citep{Lin2016}, \citep{SubramaniConstraints}). 

\section{Related Work}

Robot manipulation and path planning in the presence of task constraints are an active research area in robotics. \citet{Stilman2010}  modeled constraints as a motion constraint vector specifying permissible degrees of freedom about task frame axes indicating which coordinate motion may change. This is similar to the Pl\"{u}cker representation \citep{siciliano2008springer}. This representation is useful for planning but difficult to use for estimation and fitting geometry as some constraint models have coupled linear and angular rotations and may not have a well defined task frame attached to the motion of the body. For example, in the point on plane constraint, the rotation axis does not coincide with the point of contact on the rigid body or is even the same along the entire motion.  
\citet{Li2017}  encoded task manifolds as a Constrained Object Manipulation (COM) task for efficient path planning using a human demonstration to learn a compliance controller. \citet{havoutis2013motion} represented constraints as lower dimensional task manifolds learned through manifold learning. While this can encode complicated constraint regions, it does not allow a compact, semantic, parameterized representation of a constraint (for example, a hinge constraint is defined by its axis or rotation).
\citet{Ortenzi2016} used the null space of allowable velocities to determine constraints autonomously but this may not be suitable for a human demonstration setting.

Constraint inference has also been explored in the Programming by demonstration(PbD) context. C-Learn by \citet{perez2017c}, utilizes different grasp approaches and key-frames to determine permissible directions such as planar motion or motion along a line. CHAMP by \citet{Niekum2015}, uses a parametric model based change point detection system to determine geometric objects such as lines and arcs in recorded motions. Inferring constraints by learning the null spaces of motion has been explored by \citet{Lin2016}. All these approaches provide semantic constraint representations but can only identify simple constraints such as lines, planes and arcs. 
Using forces for constraint inference has been explored too. In \citet{SubramaniConstraints}, clustering and filtering based approaches were used to determine plane, line and arc constraint geometry. Constrained motions during multiple contacts between a polyhedral robotic tool and the environment are explored by \citet{Meeussen2008}. All these methods are limited to simple constraints and do not incorporate moment and orientation information requiring specification of tool geometry. 

Our method distinguishes itself from the above-mentioned inference methods by identifying the constraint type and estimating the parameters associated with the constraint model explicitly. Constraints such as \textit{a point on an object constrained on a plane} can be inferred without any knowledge of the location of the point of contact, something that no other prior work considers let alone can infer. The models used are compact and semantic, yet can represent constraints of multiple degrees of freedom. Our method can incorporate both force and moment information to identify constraint geometry that would otherwise be ambiguous when only kinematic information is known. 

\section{Salient Features of Approach}
This paper provides a method to infer geometric constraints in human demonstrations. Our method takes as input recorded positions, orientations, forces and moments and outputs constraints over a demonstration. The salient features of this approach are as follows:
\squishlist
\item Prior to processing, a library of constraint types is constructed. Kinematic constraints are modeled as constraint equations of the rigid body's configuration. 

\item Input to our method is a recorded task demonstration, consisting of measurements of positions, orientations, forces, and moments for an instrumented tool. The measurement can be in an arbitrary frame on the rigid tool; our methods will determine the contact/constraint parameters with respect to this frame.

\item Our approach requires the motions to be segmented into periods with a single constraint type. We use the approach of \citep{Subramani2017} to perform the segmentation.

\item For each segment, our approach first attempts to fit each constraint model in the library, estimating their corresponding parameters using nonlinear least squares regression.

\item Each sample in the segment is checked for permissible position, force and moment errors with each constraint type. Constraints are often ambiguous from kinematics alone. Our method disambiguates constraint types using force and moment information.

\item Our approach then selects the constraint through a voting process. Each sample votes for a constraint type if it satisfies the position, force and moment conditions for it.
\squishend

\begin{figure}[t]
\includegraphics[width=5.5in]{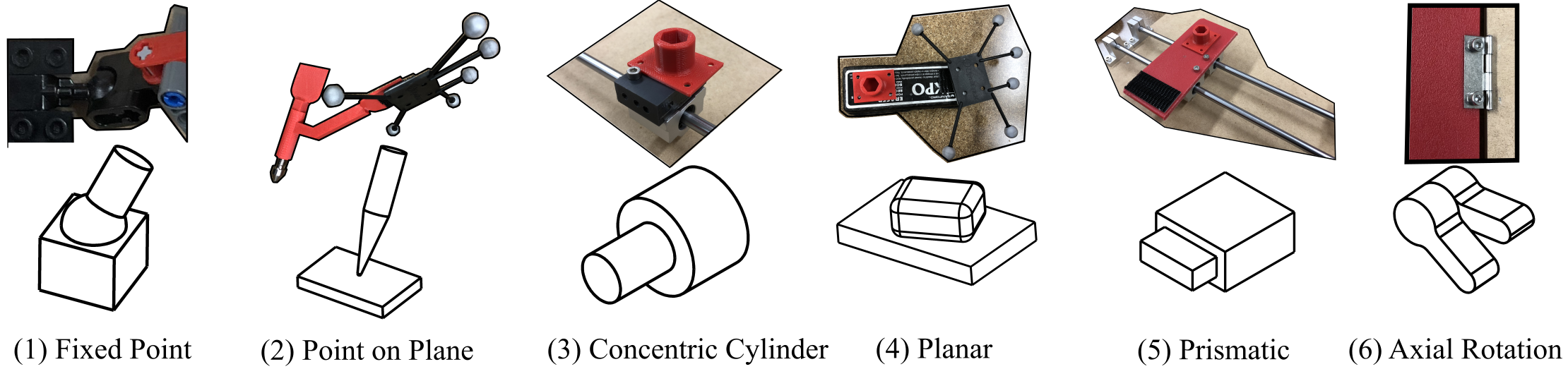}
\caption{Geometric constraints and their physical counterparts}
\vspace*{-0.2in}
\label{fig:constraints}
\end{figure}

\section{Mathematical Modeling}
When an object/robot end effector is constrained, the degrees of freedom of its rigid body motion are limited and its rigid body motion is restricted to a subset in SE(3). 
Our modeling approach models the constrained object as a rigid body. Constraint equations are mathematical relationships between defined geometry on the body (e.g., a point on the body) and defined global geometry (e.g., a plane in the environment). 
For an initial library of constraints, we have analyzed the constraints shown in figure \ref{fig:constraints}. Other constraints can be added if desired. Our derivations use a standard representation for rigid body motion, alternate representations could be used as well.

\subsection{Generalized Model of Constraints}
This section derives the relationship between the generalized constraint geometry and the permissible linear velocity, angular velocity, forces and moments on the body. First, we determine the permissible linear and angular velocities using virtual displacements. Next, we use the principle of virtual work to determine the permissible reaction forces and moments.

Consider a 6-degree of freedom rigid body of negligible inertial properties located in space through 3 translational coordinates r  and rotation coordinates represented either with a unit quaternion q $\in$   SU(2) or an orthogonal rotational matrix $\mathbf{A}(\text{q}) \in \text{SO(3)}$. 
The rigid body is constrained by $k$ constraint equations $\Phi$ : SE(3) $\Rightarrow$ $\mathbb{R}^k$ such that:
\begin{equation}\label{eq:phi}
\Phi(\text{p}) = 0
\end{equation}
where $\text{p}=(\text{r},\text{q})$. These equations ($\Phi$) represent the configurations of the rigid body. 

In order to be admissible, virtual displacements $\delta \text{p}$ (variation of \text{p}) under the constraint equations (\ref{eq:phi}) must satisfy:
\begin{equation}\label{eq:phiexp}
\delta \Phi ={{\Phi }_{\text{r}}}\delta \text{r}+{{\Phi }_{\text{q}}}\delta \text{q} = 0
\end{equation}
Where ${{\Phi }_{r}}\delta \text{q}$ and ${{\Phi }_{\text{q}}}\delta \text{q}$ are the partial derivatives of (\ref{eq:phi}). Equation (\ref{eq:phiexp}) may also be written using virtual rotation variable $\delta \pi $ \citep{haug1989computer}. The virtual rotation variable is related to the angular velocity of a rigid body. This relationship is similar to how the virtual displacement $\delta \text{p}$ is related to linear velocity. 
\begin{equation}\label{eq:virt}
\delta \Phi ={{\Phi }_{\text{r}}}\delta \text{r}+{{\Phi }_{\pi }}\delta \pi = 0
\end{equation}
The permissible reaction forces $\text{f}\in {{\mathbb{R}}^{3}}$ and reaction moments $\text{n}\in {{\mathbb{R}}^{3}}$ that the constraint applies to the constrained body must satisfy the virtual work equation \citep{lanczos2012variational} because constraint reaction forces and moments do not produce work. The principle of virtual work requires:
\begin{equation} \label{eq:vw}
\delta {{\text{r}}^{T}}\text{f}+\delta {{\pi }^{T}}\text{n}=0
\end{equation}
Combining equations (\ref{eq:virt}) and (\ref{eq:vw}):
\begin{gather}
{{\Phi }_{\text{r}}}^{T}\lambda+\text{f}=0 \label{eq:vwf}\\
\Phi _{\pi }^{T}\lambda+\text{n}=0 \label{eq:vwm}
\end{gather}
where $\lambda\in {{\mathbb{R}}^{k}}$ are the Lagrange multipliers. Equations (\ref{eq:vwf}) and (\ref{eq:vwm}) provide a relationship between the generalized constraint equations $\Phi$ and the permissible reaction forces and moments.
${{\Phi }_{\pi }}\delta \pi $ is computed from ${{\Phi }_{\text{q}}}\delta \text{q} $ and this is shown in the Appendix (Section \ref{sec:appendix}). This formulation is standard in the multi-body dynamics literature, see \citep{haug1989computer} for a review.

\section{Mathematical models of geometric constraints}\label{sec:models}
This section describes the different constraint types evaluated in the paper. 

Consider a 6-degree of freedom rigid body of negligible inertial properties located in space through 3 translational coordinates r $\in \mathbb{R}^3$  and rotation coordinates represented either with a unit quaternion q $\in$   SU(2) or an orthogonal rotational matrix $\mathbf{A}(\text{q}) \in \text{SO(3)}$. Together (r,q) form the body coordinates p. 

Each constraint type has a set of parameters that parameterize the geometry of the constraint. Let the parameters of the constraint be $\mathbf{\alpha}$, then equation $\Phi$ becomes $\Phi(\text{p},\mathbf{\alpha})$.

\subsection{Fixed point constraint:} 
The fixed point constraint represents a point on the rigid body constrained to the environment such as a ball and socket joint. 

Consider a point $\text{s*} \in \mathbb{R}^3$ in the global reference frame defined as a fixed point on the rigid body. s is a vector directed from the origin of the local reference frame of the body to the rigidly fixed point s* in the global reference frame and its local reference frame counterpart is$\overset{-}{\text{s}}$ :
\begin{equation}
\text{s*}\equiv \text{r}+\text{s}\equiv \text{r}+\textbf{A}\overset{-}{\text{s}}
\end{equation}
where r and A are the body coordinates. s* is attached to a point P $\in \mathbb{R}^3$ in the environment to create the fixed point constraint: 
\begin{equation}
\Phi \equiv \text{r}+\mathbf{A}\overset{-}{\text{s}}-\text{P}=0
\end{equation}
where $\Phi$ : SE(3) $\Rightarrow \mathbb{R}^3$. The parameters of this constraint are P and $\overset{-}{\text{s}}$ (a total of six variables). $\alpha = (\text{P},\overset{-}{\text{s}})$ This constraint removes three degrees of freedom through three constraint equations leaving three degrees of freedom of motion. 

\subsection{Point on plane constraint:} 
The point on plane constraint describes a point on the rigid body constrained to a plane in the environment such as a pencil tip (point) moving across paper (plane).

This constraint requires a representation of a plane. A plane may be generated by applying a general displacement (i.e. translation and rotation) transformation of the x-y plane which involves: 
\begin{enumerate}
\item translating the x-y coordinate plane along the z-axis
\item rotating the translated plane about the origin. 
\end{enumerate}
We represent the rotation transformation using two exponential coordinates $\text{w}=[{{w}_{x}},{{w}_{y}},0]\in \mathbb{R}^2$ corresponding to ${{e}^{{\tilde{w}}}}\in SO(3)$, the exponential map, which is equivalent to a rotation matrix with an axis of rotation in the x-y plane. Rodrigues’ rotation formula \citep{murray1994mathematical} is used to compute ${e}^{\tilde{w}}$. The third term of w is zero because rotations about the z - axis (perpendicular to the plane) do not alter the plane’s geometry. The translation is represented by $d \in \mathbb{R}$. Thus, the normal vector on this plane is represented by ${e}^{\tilde{w}}{\left[ 0\ \ 0\ \ 1 \right]}^{T}$ and the shifted origin of the x-y plane is represented by ${{e}^{{\tilde{w}}}}{{\left[ 0\ \ 0\ \ d \right]}^{T}}$.

A point P on the plane satisfies the following equation:
\begin{equation}
{{\left( {{e}^{{\tilde{w}}}}{{\left[ 0\ \ 0\ \ d \right]}^{T}}-\text{P} \right)}^{T}}{{e}^{{\tilde{w}}}}{{\left[ 0\ \ 0\ \ 1 \right]}^{T}}=0
\end{equation}
which specifies that the dot product between a vector within the plane and the plane normal is zero. If the constrained point on the rigid body is s* then the constraint equation $\Phi:SE(3) \Rightarrow \mathbb{R}^1$ is:
\begin{equation}
{{\Phi }_{1}}\equiv {{\left( {{e}^{{\tilde{w}}}}{{\left[ 0\ \ 0\ \ d \right]}^{T}}-\text{r}-\textbf{A}\bar{\text{s}} \right)}^{T}}{{e}^{{\tilde{w}}}}{{\left[ 0\ \ 0\ \ 1 \right]}^{T}}=0
\end{equation}

The parameters of this constraint are $\alpha  =  (\bar{\text{s}},d,{{w}_{x}},{{w}_{y}})$, six variables. The constraint equation removes one degree of freedom leaving 5 degrees of freedom of movement.

\subsection{Concentric cylinder constraint:} The concentric cylinder constraint is similar to the motion of a collar on a shaft where a rigid body (the collar) is permitted to translate and rotate about a fixed axis (shaft).

This constraint requires a representation of the axis of rotation.  Similar to the plane, the axis of rotation can be generated by applying a general displacement (i.e. translation and rotation) transformation of the z coordinate axis which is equivalent to:
\begin{enumerate}
\item translating the z axis in the x-y plane
\item rotating the translated axis about the origin. 
\end{enumerate}

This is represented by two exponential coordinates  $\textbf{w}=[{{w}_{x}},{{w}_{y}},0]\in \mathbb{R}^2$ and two translational coordinates $d_x , d_y \in \mathbb{R}$. The third term in \textbf{w} is zero because rotations about the z - axis produce a line that could be produced by an alternative translation motion. This defines the axis in the global reference frame. The tangent to this axis is ${{e}^{{\tilde{w}}}}{{\left[ 0\ \ 0\ \ 1 \right]}^{T}}$ and the translated origin is ${{e}^{{\tilde{w}}}}{{\left[ {{d}_{x}}~{{d}_{y}}~0 \right]}^{T}}$. ${{e}^{{\tilde{w}}}}{{\left[ 1\ \ 0\ \ 0 \right]}^{T}}$ and ${{e}^{{\tilde{w}}}}{{\left[ 0\ \ 1\ \ 0 \right]}^{T}}$  represent vectors perpendicular to this axis. 
The rigid body must only translate and rotate about this axis and this is enforced by :
\begin{enumerate}
\item constraining a point s* on the rigid body to coincide with the axis 
\item constraining vector s to be perpendicular to this axis 
\item constraining a unit vector $\bar{\text{t}}$ on the rigid body to be perpendicular to $\bar{\text{s}}$ and the axis.
\end{enumerate}
The unit vector $\bar{\text{t}}$ prevents the rigid body from rotating about $\bar{\text{s}}$. The constraint equations are:
\begin{gather}
{{\Phi }_{1}}\equiv {{\left( {{e}^{{\tilde{w}}}}{{\left[ {{d}_{x}}~{{d}_{y}}~0 \right]}^{T}}-\text{r}-\textbf{A}\bar{\text{s}} \right)}^{T}}{{e}^{{\tilde{w}}}}{{\left[ 1\ \ 0\ \ 0 \right]}^{T}}=0 \label{eq:c1}\\
{{\Phi }_{2}}\equiv {{\left( {{e}^{{\tilde{w}}}}{{\left[ {{d}_{x}}~{{d}_{y}}~0 \right]}^{T}}-\text{r}-\textbf{A}\bar{\text{s}} \right)}^{T}}{{e}^{{\tilde{w}}}}{{\left[ 0\ \ 1\ \ 0 \right]}^{T}}=0\label{eq:c2}\\
{{\Phi }_{3}}\equiv {{\left( \textbf{A}\bar{t}~ \right)}^{T}}{{e}^{{\tilde{w}}}}{{\left[ 0\ \ 0\ \ 1 \right]}^{T}}=0\label{eq:c3}\\
{{\Phi }_{4}}\equiv {{\left( \textbf{A}\bar{\text{s}}~ \right)}^{T}}{{e}^{{\tilde{w}}}}{{\left[ 0\ \ 0\ \ 1 \right]}^{T}}=0\label{eq:c4}\\
{{\Phi }_{\mathbf{p}1}}\equiv {{\bar{\text{s}}}^{T}}\bar{\text{t}}=0\label{eq:c5}
\end{gather}
Equations (\ref{eq:c1}) and (\ref{eq:c2}) enforce point $\bar{\text{s}}$ to lie on the axis because the vector between the origin of the line and the point s* must be perpendicular to ${{e}^{{\tilde{w}}}}{{\left[ 1\ \ 0\ \ 0 \right]}^{T}}$ and ${{e}^{{\tilde{w}}}}{{\left[ 0\ \ 1\ \ 0 \right]}^{T}}$ .  
Equations (\ref{eq:c3}), (\ref{eq:c4}) and (\ref{eq:c5}) enforce perpendicularity between $\bar{\text{t}}$, $\bar{\text{s}}$ and the axis. 

A solution to equations (\ref{eq:c3}) and (\ref{eq:c5}) is $\bar{\text{t}} = 0$. This solution does not provide the intent of these constraint equations which is to force perpendicularity between vectors.  To address this, we constrain $\bar{\text{t}}$ to equal a unit vector:
\begin{equation}\label{eq:c6}
{{\Phi }_{\mathbf{p}2}}\equiv {{\bar{\text{t}}}^{T}}\bar{\text{t}}-1=0
\end{equation}

Equations (\ref{eq:c1}) through (\ref{eq:c6}) represent the constraint equations $\Phi: SE(3) \Longrightarrow \mathbb{R}^6 $. 
The parameters of this constraint are $\alpha  =  (\bar{\text{t}},\bar{\text{s}},{{d}_{x}},{{d}_{y}},{{w}_{x}},{{w}_{y}})$. Parameter equations (\ref{eq:c5}) and (\ref{eq:c6}) do not apply a constraint on the body but define geometry on it and thus do not remove degrees of freedom from the body. Thus the rigid body has 2 degrees of freedom of movement.

\subsection{Planar constraint:} The planar constraint is exemplified by an eraser moving against a whiteboard. The rigid body can only rotate about a vector perpendicular to the plane, and all points within the rigid body translate parallel to this plane. We may assume the origin of the local coordinate frame on the body is contained within the plane.
\begin{equation}
{{\Phi }_{1}}\equiv {{\left( {{e}^{{\tilde{w}}}}{{\left[ 0\ \ 0\ \ d \right]}^{T}}-\text{r} \right)}^{T}}{{e}^{{\tilde{w}}}}{{\left[ 0\ \ 0\ \ 1 \right]}^{T}}=0
\end{equation}

A unit vector $\bar{\text{t}}$ (unity enforced by equation (\ref{eq:c6})) is defined to force the rigid body perpendicular to the plane: 
\begin{gather}
{{\Phi }_{2}}\equiv {{\left( \textbf{A}\bar{\text{t}}~ \right)}^{T}}{{e}^{{\tilde{w}}}}{{\left[ 1\ \ 0\ \ 0 \right]}^{T}}=0\label{eq:p1}\\
{{\Phi }_{3}}\equiv {{\left( \textbf{A}\bar{\text{t}}~ \right)}^{T}}{{e}^{{\tilde{w}}}}{{\left[ 0\ \ 1\ \ 0 \right]}^{T}}=0\label{eq:p2}
\end{gather}

The parameters of this constraint are $\alpha  =  (\bar{\text{t}},d,{{w}_{x}},{{w}_{y}})$, a total of 6 variables. Thus, the rigid body has 3 degrees of freedom of movement.

\subsection{Prismatic constraint:}
The prismatic constraint represents translational motion in one direction. It is similar to pulling out a drawer. All points on the rigid body translate identically. We assume the origin of the local coordinate frame of the body is contained within this line/axis:
\begin{gather}
{{\Phi }_{1}}\equiv {{\left( {{e}^{{\tilde{w}}}}{{\left[ {{d}_{x}}~{{d}_{y}}~0 \right]}^{T}}-\text{r} \right)}^{T}}{{e}^{{\tilde{w}}}}{{\left[ 1\ \ 0\ \ 0 \right]}^{T}}=0\label{eq:l1}\\
{{\Phi }_{2}}\equiv {{\left( {{e}^{{\tilde{w}}}}{{\left[ {{d}_{x}}~{{d}_{y}}~0 \right]}^{T}}-\text{r} \right)}^{T}}{{e}^{{\tilde{w}}}}{{\left[ 0\ \ 1\ \ 0 \right]}^{T}}=0\label{eq:l2}\\
\end{gather}

The orientation of the rigid body is fixed using equations (\ref{eq:c5}), (\ref{eq:c6}) and the following: 
\begin{gather}
{{\Phi }_{3}}\equiv {{(\textbf{A}\bar{\text{s}})}^{T}}{{e}^{{\tilde{w}}}}{{\left[ 1\ \ 0\ \ 0 \right]}^{T}}=0\label{eq:l3}\\
{{\Phi }_{4}}\equiv {{(\textbf{A}\bar{\text{s}})}^{T}}{{e}^{{\tilde{w}}}}{{\left[ 0\ \ 1\ \ 0 \right]}^{T}}=0\label{eq:l4}\\
{{\Phi }_{5}}\equiv {{(\textbf{A}\bar{\text{t}})}^{T}}{{e}^{{\tilde{w}}}}{{\left[ 1\ \ 0\ \ 0 \right]}^{T}}=0\label{eq:l5}\\
{{\Phi }_{\mathbf{p}3}}\equiv {{\bar{\text{s}}}^{T}}\bar{\text{s}}-1=0\label{eq:l6}
\end{gather}

where $\bar{\text{s}}$  and $\bar{\text{t}}$  are unit vectors fixed on the rigid body. Equations (\ref{eq:l3}) through (\ref{eq:l6}) and (\ref{eq:c6}) prevent the body from rotating. This rigid body has one degree of freedom of motion. The constraint parameters are: $\alpha  =  (\bar{\text{t}},\bar{\text{s}},{{d}_{x}},{{d}_{y}},{{w}_{x}},{{w}_{y}})$.

\subsection{Axial rotation constraint:} The axial rotation constraint is similar to a door knob or a hinged door, all points on the rigid body rotate about an axis and translations are not permitted.

Consider a point on the rigid body s* that is on the axis of rotation and in the plane perpendicular to the axis containing the coordinate frame origin. A rigid point on the axis of rotation defined as $({{d}_{x}},{{d}_{y}},{{d}_{z}})\in {\mathbb{R}^{3}}$ constrains the point s*. The vector  $\bar{\text{s}}$ is perpendicular to this axis. The rigid body must not rotate about vector $\bar{\text{s}}$ so unit vector $\bar{\text{t}}$ is introduced to enforce this condition. The constraint equations are (\ref{eq:c5}), (\ref{eq:c6}) and the following:
\begin{gather}
{{\Phi }_{1}}\equiv ~\text{r}+\mathbf{A}\bar{\text{s}}-{{e}^{{\tilde{w}}}}{{[{{d}_{x}},{{d}_{y}},{{d}_{z}}]}^{T}}=0\\
{{\Phi }_{2}}\equiv {{\left( \textbf{A}\bar{\text{t}}~ \right)}^{T}}{{e}^{{\tilde{w}}}}{{\left[ 0\ \ 0\ \ 1 \right]}^{T}}=0\\
{{\Phi }_{3}}\equiv {{\left( \textbf{A}\bar{\text{s}}~ \right)}^{T}}{{e}^{{\tilde{w}}}}{{\left[ 0\ \ 0\ \ 1 \right]}^{T}}=0
\end{gather}

Constraint parameters are $\alpha  =  (\bar{\text{t}},\bar{\text{s}},{{d}_{x}},{{d}_{y}},{{d}_{z}},{{w}_{x}},{{w}_{y}})$. This constraint has 1 degree of freedom of movement.

\section{Inference Approach}\label{sec:approach}
Our algorithm takes as input $n$ samples of the motion containing positions, orientations, linear velocity, angular velocity, forces and moments. As output, it provides a constraint model and its corresponding constraint parameters, $\mathbf{\alpha}$.  
Our method requires the demonstration to be segmented such that each period contains a single constraint. We perform this segmentation using the method of \citep{Subramani2017}. The segmentation allows us to assume that all samples in a segment are part of the constraint.
 The steps of our approach are as follows:
\subsection{Fitting Geometric Models to Kinematic Information}
The first step is to fit kinematic information to all of our defined constraint models.
% and determine the parameters associated with each constraint type. 
Equations (\ref{eq:phi}) and (\ref{eq:phiexp}) are used as the regression function. The virtual displacements $\delta p $ and $\delta \pi $ are replaced with linear velocity v and angular velocity $\omega$. To simplify notation, equation (\ref{eq:phi}) and (\ref{eq:phiexp}) are represented as $\Phi$ and $\delta \Phi$ respectively.
Consider the $n$ samples used for the model fit. These samples must satisfy $\Phi$ and $\delta \Phi$. The kinematic fit estimates model parameters $\mathbf{\alpha}$ using known variables p and v by inserting these values in $\Phi$ and $\delta \Phi$ at every sample and performing a least squares regression over all samples: 
\begin{equation}
\alpha =\underset{\alpha }{\mathop{\text{argmin}}}\sum\limits_{1}^{n}{\left( {\Phi }_{n}{(\text{p},\alpha )}^{T} {\Phi }_{n}(\text{p},\alpha) + \delta{{\Phi}_{n}}(\text{p},\text{v},\omega ,\alpha )^{T}\delta{{\Phi }_{n}}(\text{p,v},\omega ,\alpha ) \right)}
\end{equation}

The least squares regression was performed using Broyden-Fletcher-Goldfarb-Shanno algorithm (BFGS) \citep{Fletcher2013}. 

\subsection{Evaluating the fit quality with Kinematic Information}
Once the parameters for each constraint are identified, the best model is selected. A na\"{i}ve approach would be to pick the model with the least fit error. However this approach will fail as the equations are not comparable and the units are not the same. 
Our approach is to eliminate models that do not agree with the data. Each model is evaluated independently. A kinematic error criterion (shown in Table \ref{table:kec}) uses kinematic information to evaluate position and orientation errors for each sample in the data. 

A threshold is used to identify samples that agree well with the model. This threshold may be set considering the scale of the constraint motion (e.g. opening large a room door vs turning a small knob) and instrumentation accuracy.
%A threshold is determined to identify samples that agree well with the model based on the scale of the constraint motion and measurement errors of the position and orientation data.
This threshold generates a Boolean list $L_k$ corresponding to the permissible samples using the kinematic error criterion. The units of the kinematic error criteria are distance. 

\begin{table}[b]
	\centering
	\caption{Kinematic error criteria for each constraint}
	\label{table:kec}
	\begin{tabular}{|l|l|}
	\hline
	Prismatic Motion	&  Distance between estimated line and coordinate frame origin\\
	Axial Rotation	&  Distance between the rotation point on rigid body and axis of rotation\\
	Planar Motion	&  Distance between the rigid body frame origin and the plane\\
	Fixed Point	&  Distance between the global point P and rigid body point s*\\
	Concentric Cylinder & Distance between rotation point on body and estimated axis of rotation \\
	Point on Plane & Distance between point on rigid body and estimated plane\\
	\hline
	\end{tabular}
\end{table}

\subsection{Evaluating fit quality using Force and Moment information}
In many cases the model may still be ambiguous after applying the kinematic error criterion described above. For example, a real fixed point constrained motion will have very small kinematic errors for both the \textit{point on plane} and the \textit{fixed point} constraints because the \textit{point on plane} is the more general model. In this case the plane geometry of the \textit{point on plane constraint} is not well defined as different planes could satisfy the kinematic information. In other situations it may still be well defined. For instance, the motion of opening a door, the \textit{axial rotation constraint}, may also be a permissible \textit{planar constraint}; both the geometry of the \textit{axial rotation} and \textit{planar} constraints are well defined for the prescribe kinematic motion. This ambiguity can be addressed by considering reaction and friction forces(moments) caused by the constraint.
However, along with the reaction and friction forces, the measured forces contain inertial and gravitational forces. Inertial properties of the constrained object are not compensated but they may be considered negligible when the accelerations in the demonstration are small. 
Gravitational forces of the constrained object are not compensated but they may be considered negligible when constrained motion of the object is perpendicular to the direction of gravity or the object moved is light. 
This is the case in many real world situations such as pulling out a drawer where the motion of the draw is perpendicular to gravity or erasing on a whiteboard where the weight of the dry eraser is small.

We incorporate force information by determining whether the measured reaction forces and friction forces are consistent with the identified constraint. 
If the identified constraint is correct, then ideally, the measured forces would be equal to the sum of the reaction and friction forces (assuming negligible inertial and gravitational effects).
Equations \ref{eq:vwf} and \ref{eq:vwm} provide the permissible reaction forces and moments. To determine the reaction forces and moments, the Lagrange multipliers $\lambda$ must be estimated. Using the measured forces and moments, the Lagrange multipliers are solved at every sample using least squares optimization:

\begin{equation}\label{eq:lag}
\lambda=\underset{\lambda}{\mathop{\arg\min}}\left(
{{\left( \Phi _{r}^{T}\lambda+\text{f} \right)}^{T}}
\left( \Phi _{r}^{T}\lambda\!\!\text{ }+\text{f} \right) 
+{{\left( \Phi _{\pi }^{T}\lambda+\text{n} \right)}^{T}}\left( \Phi _{\pi }^{T}\lambda+\text{n} \right) \right) 
\end{equation}
		
The estimated reaction forces $\text{f}_\text{react}$ and estimated reaction moments $\text{n}_\text{react}$ are computed as follows: 
\begin{gather}
{{\text{f}}_{\text{react}}}=\Phi _{r}^{T}\lambda\\
{{\text{n}}_{\text{react}}}=\Phi _{\pi}^{T}\lambda
\end{gather}
The residual forces and moments after removing reaction forces and moments are:  
\begin{gather}
	\text{f}_\text{residual} = \text{f} - \text{f}_\text{react}\\
\text{n}_\text{residual} = \text{n} - \text{n}_\text{react}
\end{gather}

The residuals still contain friction. Friction forces $f_\mu$ and moments $n_\mu$ are directed along the direction of motion. $\hat{v}$ and $\hat{\omega}$ are unit vectors of velocity and angular velocity respectively. 
\begin{gather}
{\text{f}}_{\mu }=\text{f}_{\text{residual}}^{T}\hat{v}v\\
{\text{n}}_{\mu }=\text{n}_{\text{residual}}^{T}\hat{\omega}\omega
\end{gather}
Finally, the force and moment error criteria are: 
\begin{gather}
{{\text{f}}_{\text{error}}}={{\left\| \text{f}-{{\text{f}}_{\text{react}}}-{{\text{f}}_{\mu }} \right\|}_{2}}\\
{{\text{n}}_{\text{error}}}={{\left\| \text{n}-{{\text{n}}_{\text{react}}}-{{\text{n}}_{\mu }} \right\|}_{2}}
\end{gather}
It is important that friction forces and moments are removed after the reaction forces and moments are removed. It is possible to have reaction forces and moments in the direction of motion. This is because forces and moments in the virtual work equation are coupled and work done by each may cancel out the other: 
\begin{equation}\label{eq:couple}
{{v}^{T}}{{\text{f}}_{\text{react}}}+{{\omega }^{T}}{{\text{n}}_{\text{react}}}=0
\end{equation}

\begin{table}[]
\centering
\caption{Kinematic error criteria for each constraint}
\label{table:thresholds}
\begin{tabular}{ |l || c c c |}
\hline
  \textbf{Constraint Type} & \textbf{Position Threshold (m)} & \textbf{Force Threshold (N)} & \textbf{Moment Threshold (Nm)} \\
\hline
\hline
  Point on plane & 0.0005 & .05 & 0.5 \\
  Fixed point & 0.0001 & 1.0 & 0.2 \\
  Concentric cylinder & 0.002 & 0.02 & 0.01 \\
  Planar & 0.001 & 1. & 0.02 \\
  Prismatic & 0.001 & 0.1 & 0.001 \\
  Axial rotation & 0.01 & 0.02 & 0.2\\
  \hline
\end{tabular}
\end{table}

Similar to the kinematic error criterion, the force and moment error criteria are determined through a threshold of $\text{f}_\text{error}$ and $\text{n}_\text{error}$ and a boolean list $L_f$ and $L_n$ are determined. This threshold depends on the accuracy of the measured data and the size of the constraints involved. The threshold for the moments depends on the distance of the applied forces from the local reference frame, as a large distance would result in large moments. In practice, applying the thresholds is fast and can be done interactively; the easiest way to determine them is to perform a few experiments with the instrumentation and available constraint models. (The numerical values of the thresholds for the experiments in the following sections are provided in table \ref{table:thresholds}.)

\subsection{Selecting the constraint model}

Each model has corresponding kinematic, force and moment boolean lists $L_p$, $L_f$ and $L_n$ describing which samples fit the model. The intersection of these lists provide the eligible samples for each constraint type. The constraint with the most eligible samples is selected.

\begin{figure}[b]
	\includegraphics[width=5.5in]{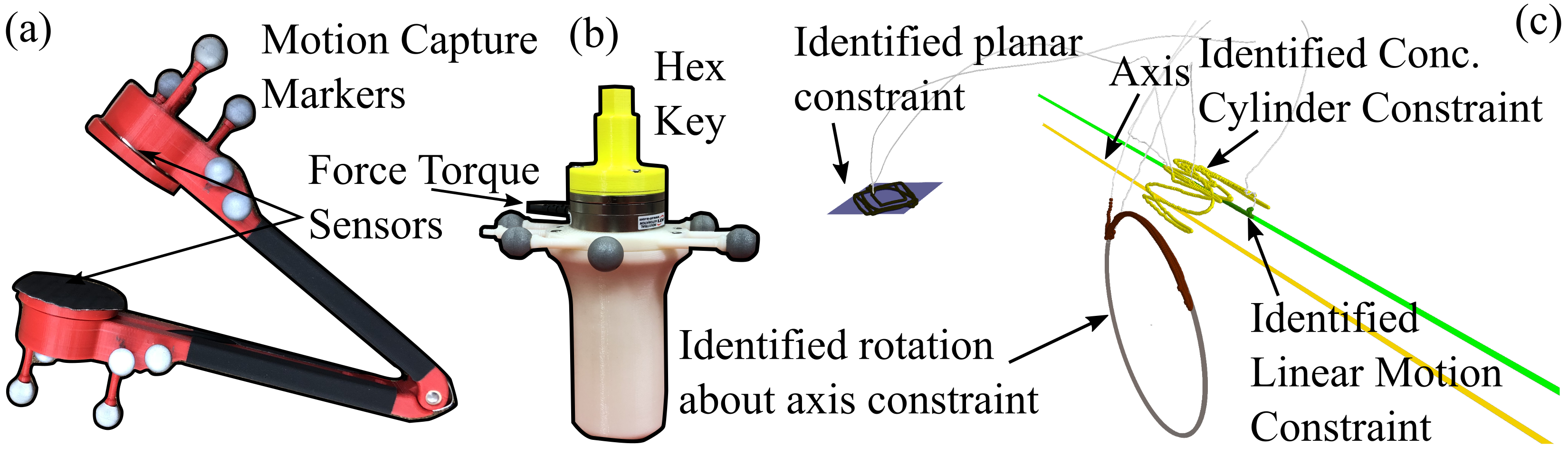}
	\caption{(a) Instrumented tongs, (b) Constraint Sabre, (c) Inferred constraint motions. The motions are correctly segmented into free space motion (grey) and constrained motion (colored).}

\label{fig:tongs}
\vspace{-0.0in}
\end{figure}

\section{Experimental Evaluations}
The performance of our constraint inference system was evaluated using a custom testbed containing the constraints described in section \ref{sec:models}. Two different hand held tools, the \textit{Instrumented tongs} and the \emph{Constraint Sabre}, shown in figure \ref{fig:tongs}, were used to collect measurements from constraint interactions in human demonstrations. 

The \emph{Instrumented tongs} simulate a robot gripper and are used to manipulate constrained objects. The tongs are tracked using an Optitrack motion capture system and applied forces and moments were measured using two enclosed ATIMini40 force torque sensors. The tongs do not provide a rigid attachment to constrained objects, and thus, the motions of the tongs cannot be considered as the motions of the constrained object, requiring markers to be placed on the object. The force-torque sensors' measurements are transformed to the object's frame to calculate the applied forces and moments on the object. 

In the case of the \emph{Constraint Sabre}, the tool attaches rigidly to a constrained object using a \emph(hex key), enabling the use of the constraint sabre kinematic measurements as a proxy for the motion of the constrained object. Motion capture markers measure the \emph{Constraint Sabre}'s position and orientation in space and a force torque sensor measures the applied forces and moments. Different tools may be attached to the \emph{Constraint sabre} to interact with constraints in the environment similar to different robot end effector tool attachments.

In section \ref{sec:approach}, the method assumes that samples considered contain only one active constraint. This is not a practical assumption, as a typical demonstration may have multiple constraints separated by free space motion. The force action recognition method of \citep{Subramani2017} was used to identify constrained motion segments between free space motion. 
%For the interested reader, experimental evaluations with a different hand held tool are provided in the supplement. 
Figure \ref{fig:tongs} shows an example of a typical demonstration and its visualization. An expert uses similar plots to determine classification accuracy.

\subsection{Inferring the point on plane constraint in detail}
Determining the constraint parameters for the \textit{the point on plane} constraint is shown here(shown in figure \ref{fig:teaser}). 
The \textit{point on plane} constraint was demonstrated by moving a custom made steel tipped stylus against a plane. 
Motion capture markers, attached to the stylus, measure its rigid body motion. A demonstrator uses the instrumented tongs to grasp the stylus and move it against the plane. 
Our method estimates both the plane geometry and the location of the point of contact on the stylus from a single demonstrated motion within the tolerance of the motion capture system. 
%Note that the location of the point of contact on the stylus is also estimated by our method. 
Any stylus (rigid body) could be used as long as measurements of the rigid body motion are available.

\subsection{Classification and fit accuracy} \label{sec:multiple}
Classification and fit accuracy were evaluated with both the \emph{Instrumented Tongs} and the \emph{Constraint Sabre} but shown as two separate experiments. 

The instrumented tongs were used to interact with 5 different constraint types. Two demonstrators, one of which was not an author of the paper, performed approximately 10 trials of each of the five constraint types (98 trials total, two trials were removed because of corruption in the motion capture data). During each trial the testbed was repositioned to provide variability in the constraint positions. 

\begin{table}[]
\centering
\caption{Classification and fit statistics for the Instrumented Tongs}
\label{table:accuracy}
\begin{tabular}{|l||lllll|} % Caution: virtical pipe line '|' and L 'l' look the same
\hline
Constraint & Classification Accuracy  &  Fit Error Mean & S.D. & Min & Max\\ \hline \hline
Prismatic & 100\% &0.000104 & 1.99e-05  & 6.58e-05  & 0.00014\\
Axial Rotation & 100\%& 0.00106 & 0.000247 & 0.000449 & 0.00148\\
Fixed Point & 75\% & 0.000771 & 0.000177  & 0.000573 & 0.00135\\
Point on Plane & 100\%& 0.000166 & 4.26e-05 & 0.000110 & 0.000248 \\
Planar Motion & 100\%&0.000894 & 0.000445 & 0.000440 & 0.00249 \\ \hline
\end{tabular}
\end{table}

%The constraint inference was 100\% successful for classifying all constraint types except for the \textit{fixed point constraint} where the classification accuracy was 75\%
The classification accuracy and corresponding fit accuracy calculated using the Kinematic error criterion (assuming constraint is inferred correctly) is presented in table \ref{table:accuracy}. 
The \textit{fixed point constraint} was falsely predicted as a \textit{rotational constraint} (5\%) and a \textit{point on plane constraint} (20\%) because the reaction forces and moments are coupled in all three constraints (See equation (\ref{eq:couple})). Equation (\ref{eq:lag}) is not reliable when large moments are measured as large moments bias the estimated Lagrange multipliers. Without force and moment thresholding the overall accuracy was 57\%. Thus, the use of force and moment information is useful for distinguishing between constraint types in ambiguous situations.
%The loss in performance when position information is only used can be seen from the following example. 

A similar experiment was conducted with the \emph{Constraint Sabre}. 
Using the constraint sabre, two demonstrators, one of whom is not an author on this paper, performed 10 and 13 constraint interactions over the \emph{linear}, \emph{axial rotation}, \emph{concentric cylinder} and \emph{planar motion} constraints. 

The complete set of constraint interactions were performed over the same demonstration as shown in Figure \ref{fig:tongs}. Free space motion occurred between distinct constraint interactions. The overall prediction accuracy for the linear, axial rotation, concentric cylinder and planar constraints was 87\%, 96\%, 91\% and 100\% respectively. 

The \emph{linear} and \emph{rotational} constraints are occasionally misclassified as a \emph{planar} constraint(9\% and 4\% respectively) likely because the planar constraint is a more general model which often results in a better fit to the measured data. If small forces and moments are applied during the demonstration, the force and moment criteria are not useful and thus the planar constraint is mistakenly estimated. 

Sometimes, the \emph{concentric cylinder} constraint is incorrectly identified as an \emph{axial rotation} constraint (5\%) when no translation occurred or due to the low tolerance set on the kinematic information criterion for the axial rotation constraint. The low tolerance was required to counteract the mechanical play between the hexagonal key and the physical constraint. 

The overall fit accuracy was less when compared to the instrumented tongs because of the mechanical play in the hex key mate. If the rigid body motions of the constrained objects were measured directly, the fit and classification accuracies would be similar to that of the instrumented tongs.

\begin{figure}[h]
	\includegraphics[width=5.5in]{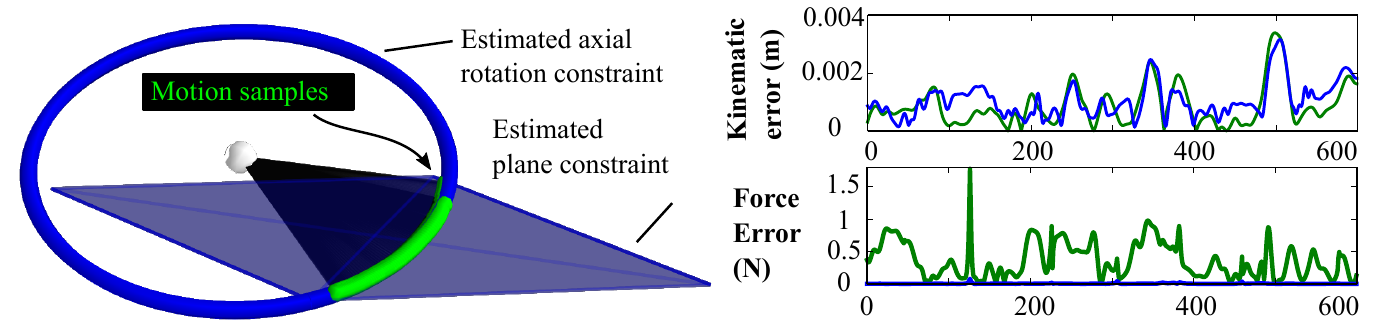}
	\caption{Left - Shows recorded motion interacting with an axial rotation constraint. Both planar and axial rotation constraint geometry fit the recorded motions. Right - While position errors for both the planar (green) and axial rotation (blue) models are similar, the force errors are significant for the planar one and can be used to determine the axial rotation model as the correct constraint model.}
\label{fig:hinge}
\vspace*{-0.2in}
\end{figure}

\subsection{Significance of force and moment information} 
The same motion may fit to different constraint models.
Figure \ref{fig:hinge} shows recorded motion interacting with an \textit{axial rotation constraint}. While position errors for both the \textit{planar} and \textit{axial rotation} models are similar as both fit the motion well, the force errors are significant for the planar one and can be used to determine the axial rotation model as the correct constraint model. Note that position errors are low for \textit{point on plane} and \textit{axial rotation} constraints too but are not shown here.
Force and moment information is useful to distinguish between constraint types that are ambiguous when only considering kinematic information. While one may argue to always pick the most constrained model and ignore force and moment information, this does not work for all situations, for example, a 2DOF \textit{concentric cylinder constraint} will fit a 3DOF \textit{point contact constraint} motion better but is a more restrictive model(lesser degrees of freedom). Including force and moment information is not a heuristic as it exploits the physics of the interaction.

\subsection{Significance of quality of information}
\begin{figure}[h]
	\includegraphics[width=5.5in]{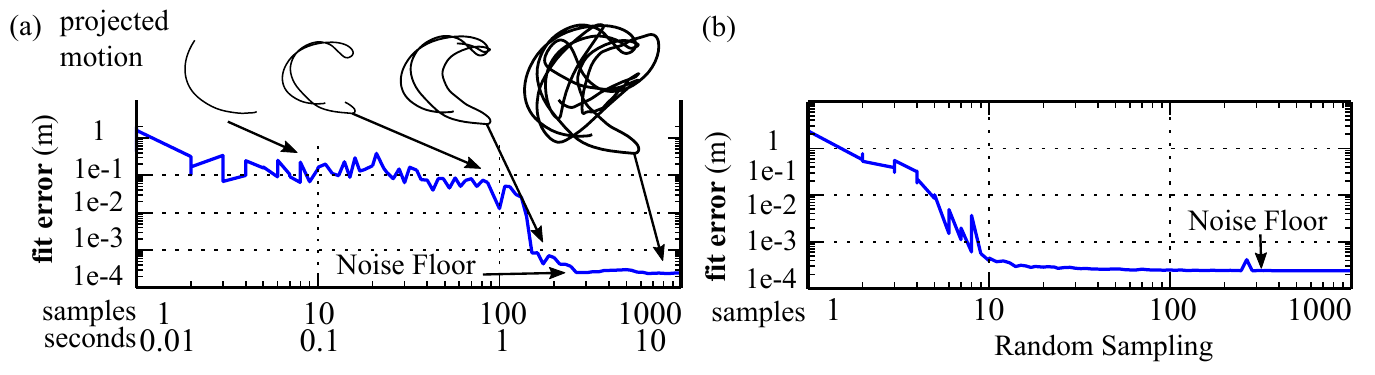}
	\caption{The \textit{point on plane constraint} performed with the instrumented tongs. The recorded trajectory is similar to the one shown in figure \ref{fig:teaser}. 
		Left - shows the fit error(m) of the point on plane constraint as a function of the number of contiguous samples used during fitting. Projected images of the stylus motion are shown indicating the amount of information over the demonstration. In this particular example, only after reaching 1 second of motion does the fit error drop to an acceptable value. Right - Samples are selected randomly from the entire demonstration and fit. The fit error drops with an increase in the number of samples. Randomly selected samples are usually distinct and provide richer information than an equal number of contiguous samples.}
\label{fig:sampling}

\end{figure}

Constraints that have many degrees of freedom such as the \textit{point on plane} constraint require many distinct samples to estimate reliable parameters. For example, simply moving the stylus in a straight line without changing its orientation in space would not provide enough information to determine the correct parameters such as the location of the plane. In a human demonstration, samples close in time are similar. Fitting this constraint with few contiguous samples in time may not provide as good a fit as sampling them randomly from the entire demonstration. Figure \ref{fig:sampling} shows this in detail.

\section{Conclusion}
This paper demonstrates a method to identify geometric constraints of various degrees of freedom even in the absence of tool geometry. It describes a method to model these constraints, fit motions to these models and uses force and moment information to distinguish between constraints in ambiguous situations. The estimated models are semantic and may be parsed by both human and machine. 

However, the proposed method does not estimate inertial and gravitational effects which in certain cases are significant, for example, interacting with constrained exercise equipment. It also does not incorporate force and moment information in the fitting procedure which may improve performance.
Finally, it would be interesting to see this method in a fully integrated teaching by demonstration system and we consider this as an interesting avenue for future work.
\section{Appendix - Kinematic Identities}\label{sec:appendix}
\subsection{Relationship between $\pi$ and q}
\begin{equation}
{{\Phi}_{\pi }}=~{{\Phi}_{\text{q}}}\frac{1}{2}{{\textbf{G}}^{T}}{{\textbf{A}}^{\text{T}}}
\end{equation}
Quaternion $\text{q}=\left[ {{e}_{0}}~~{{\text{e}}^{T}} \right]^T$ where ${{e}_{0}}={{q}_{0}}$ and $\text{e}={{\left[ {{q}_{1}}~{{q}_{2}}~{{q}_{3}} \right]}^{T}}$.
\begin{equation}
\textbf{A}=:\left( e_{0}^{2}-{{\text{e}}^{T}}\text{e} \right)I+2\text{e}{{\text{e}}^{T}}+2{{e}_{0}}\tilde{\text{e}}
\end{equation}
\begin{equation}
\textbf{G}=:~\left[ \begin{matrix}
-\text{e} & -\tilde{\text{e}}+{{e}_{0}}\textbf{I}  \\
\end{matrix} \right]
\end{equation}
where 
\begin{equation}
\tilde{\text{e}}=\left[ \begin{matrix}
0 & -{{e}_{z}} & {{e}_{y}}  \\
{{e}_{z}} & 0 & -{{e}_{x}}  \\
-{{e}_{y}} & {{e}_{x}} & 0  \\
\end{matrix} \right]
\end{equation}

and \textbf{I} is the identity matrix

%===============================================================================
\begin{comment}
\section{Citations}
\label{sec:citations}
\end{comment}
\bibliography{refs}
%===============================================================================

%===============================================================================

% The maximum paper length is 8 pages excluding references and acknowledgements, and 10 pages including references and acknowledgements

\clearpage
% The acknowledgments are automatically included only in the final version of the paper.
\acknowledgments{This work was
supported in part by NSF award 1830242 and the University of Wisconsin-
Madison Office of the Vice Chancellor for Research and Graduate Education
with funding from the Wisconsin Alumni Research Foundation.}

%===============================================================================

  % .bib

\end{document}